\documentclass[letterpaper]{article} % DO NOT CHANGE THIS
\usepackage{aaai2027}  % DO NOT CHANGE THIS
\usepackage{times}  % DO NOT CHANGE THIS
\usepackage{helvet}  % DO NOT CHANGE THIS
\usepackage{courier}  % DO NOT CHANGE THIS
\usepackage[hyphens]{url}  % DO NOT CHANGE THIS
\usepackage{graphicx} % DO NOT CHANGE THIS
\usepackage{natbib}  % DO NOT CHANGE THIS AND DO NOT ADD ANY OPTIONS TO IT
\usepackage{caption} % DO NOT CHANGE THIS AND DO NOT ADD ANY OPTIONS TO IT
\usepackage{algorithm}
\usepackage{algorithmic}
\usepackage{amsmath}
\usepackage{color}
\usepackage{multirow}
\usepackage{booktabs} 
\usepackage{amssymb}
\usepackage{xcolor}         % colors

\usepackage{graphicx}
\usepackage{amsmath}
\usepackage{color}
\usepackage{multirow}
\usepackage{algorithm}
\usepackage{algorithmic}
\usepackage{threeparttable}
\usepackage[capitalize,noabbrev]{cleveref}
\crefname{theorem}{Thm.}{Thm.}
\crefname{lemma}{Lem.}{Lem.}
\crefname{figure}{Fig.}{Fig.}
\crefname{table}{Tab.}{Tab.}
\crefname{algorithm}{Algorithm}{Algorithm}
\crefname{section}{Sec.}{Sec.}
\crefname{appendix}{Appendix}{Appendix}
\crefname{equation}{Eq.}{Eq.}

\definecolor{Red}{RGB}{192, 0, 0}
\definecolor{Blue}{RGB}{0, 0, 192}
\definecolor{Green}{RGB}{0, 192, 0}

\usepackage{newfloat}
\usepackage{listings}
\DeclareCaptionStyle{ruled}{labelfont=normalfont,labelsep=colon,strut=off} % DO NOT CHANGE THIS
\floatstyle{ruled}
\newfloat{listing}{tb}{lst}{}
\floatname{listing}{Listing}
\title{{UniDiffFusion: A Unified Diffusion Framework for Multi-Task and Degradation-Robust Image Fusion}
\author{Xingxin Xu$^{1}$, Siqi Zhao$^{2}$, Xin Li$^{3}$, Xinjie Yao$^{2}$, Yiming Sun$^{3}$\thanks{Corresponding author.}, Pengfei Zhu$^{2,3,4}$}}
\affiliations{
    \textsuperscript{\rm 1}School of New Media and Communication, Tianjin University\\
 \textsuperscript{\rm 2} School of Artificial Intelligence, Tianjin University\\
  \textsuperscript{\rm 3} School of Automation, Southeast University  \\
   \textsuperscript{\rm 4} Xiong'an Guochuang Lantian Technology Co., Ltd. \\
   \{xuxingxin, siqizhao, huqinghua, zhupengfei\}@tju.edu.cn, \{lixin\_seu, sunyiming\}@seu.edu.cn
}

\usepackage{bibentry}
\begin{document}

\maketitle

\begin{abstract}
General image fusion aims to integrate complementary information from multiple source images, but existing methods often rely on task-specific models and struggle to maintain robust performance under diverse degradation conditions. In this paper, we propose UniDiffFusion, a unified diffusion framework for multi-task and degradation-robust image fusion. UniDiffFusion leverages the strong generative prior of a pretrained diffusion model to establish a shared fusion backbone across heterogeneous fusion tasks, while introducing task- and degradation-aware conditional adaptation to accommodate their distinct information-selection requirements. Specifically, we employ task prompt modulation to progressively adapt the shared diffusion representations to different fusion objectives, and develop a degradation prompt router to dynamically retrieve degradation-aware priors and restore corrupted source features before fusion. Furthermore, an application prompt bank is introduced to incorporate task-oriented semantic guidance for downstream applications, such as object detection and semantic segmentation, without altering the shared fusion and restoration pathways. The proposed framework is trained in a progressive manner to decouple fusion learning, degradation-aware restoration, and application-specific adaptation, thereby reducing interference among heterogeneous objectives. Extensive experiments on visible-infrared, multi-exposure, and multi-focus image fusion demonstrate that UniDiffFusion achieves superior fusion quality and robustness under both clean and degraded conditions. Moreover, UniDiffFusion consistently improves downstream detection and semantic segmentation performance, demonstrating its effectiveness as a unified diffusion framework for both perceptual fusion and task-oriented vision.

\end{abstract}
\section{Introduction}

Image fusion aims to synthesize complementary information from multiple source images into a single informative and visually faithful output~\cite{zhang2021image, zhang2023visible}. It has been widely studied in diverse scenarios, including infrared-visible image fusion (VIF)~\cite{ma2019infrared, ma2023infrared}, multi-exposure fusion (MEF)~\cite{xu2020mef, ma2017robust}, and multi-focus fusion (MFF)~\cite{liu2017multi, liu2020multi}. By preserving the complementary advantages of different inputs, fusion improves visual quality and provides informative inputs for downstream applications such as object detection, semantic segmentation, and medical diagnosis~\cite{sun2023visible, ma2023infrared, zhang2023visible}, semantic segmentation~\cite{liu2023multi, zhang2024mrfs}, and medical diagnosis~\cite{james2014medical, tirupal2021multimodal}. These increasingly diverse deployment scenarios demand fusion models that are not only visually effective but also robust to degradations and adaptable to downstream perception.

\begin{figure}[t]
\centering
\includegraphics[width=1\linewidth]{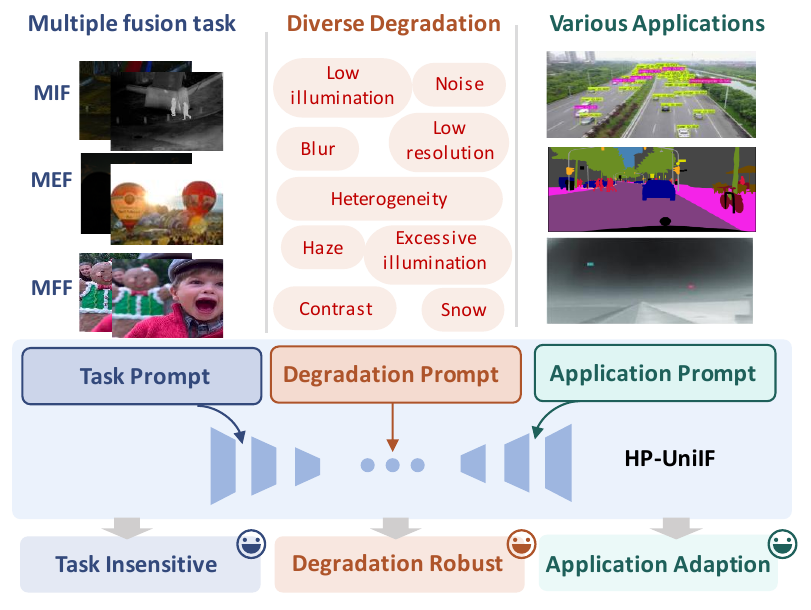}
\caption{
Motivation of HP-UniIF under the multi-task, multi-degradation, and multi-application setting. Existing task-specific, restoration-enhanced, and application-oriented paradigms address only partial requirements. In contrast, HP-UniIF coordinates task, degradation, and application prompts within a shared diffusion prior, enabling unified adaptation across heterogeneous fusion tasks, diverse degradation, and various downstream applications.
}
\label{fig:motivation}
\end{figure}

% Existing studies can be broadly grouped according to whether they prioritize perceptual fusion quality, general image ufsion ~\cite{li2021rfn, xu2020u2fusion}, degradation robustness, or downstream task utility.
Recent deep fusion models have substantially improved performance across different scenarios~\cite{li2021rfn, xu2020u2fusion}. One line of work focuses on high-quality fusion under relatively clean conditions~\cite{li2018densefuse, ma2019fusiongan}. Another line addresses adverse environments by incorporating restoration mechanisms to mitigate degradations before or during fusion~\cite{tang2022piafusion, xu2026dream}. A third line optimizes fused representations for downstream tasks, recognizing that fused images are often intermediate inputs rather than final outputs~\cite{liu2022target, tang2022image}. Nevertheless, existing methods typically address fusion quality, degradation robustness, and downstream adaptability as separate problems, relying on specialized models for individual settings and lacking a unified framework capable of jointly accommodating these requirements.

In practical deployment, heterogeneous fusion tasks, degraded inputs, and application-specific semantic requirements may coexist within the same system. We refer to the requirement of supporting multiple fusion tasks, multiple degradation conditions, and various downstream applications with a shared fusion backbone as the $\mathbf{M}^3$ setting. The central difficulty is not merely parameter sharing. Fusion, restoration, and application supervision emphasize different properties of the representation, and directly optimizing them together may cause their gradients and conditioning signals to interfere. Therefore, a unified model should preserve a strong shared visual prior while introducing the three types of conditions through structurally separated adaptation paths.
Addressing this challenge therefore calls for a model that combines strong image priors with sufficient structural flexibility to introduce task-, degradation-, and application-specific conditions.

Pretrained diffusion models provide hierarchical image representations and strong generative priors, making them a promising backbone for this purpose. However, existing diffusion-based fusion methods mainly use diffusion models as high-quality generation priors, without coordinating fusion-task, degradation, and application conditions within a single framework. Motivated by this observation, we propose HP-UniIF, a unified generative framework that jointly addresses the $\mathbf{M}^3$ challenge by supporting diverse fusion tasks, robust restoration under complex environments, and adaptation to multiple downstream applications using the diffusion priors. Concretely, HP-UniIF introduces a hierarchical prompt-learning strategy that organizes heterogeneous conditions across three complementary functional stages of the diffusion framework. Task prompt modulation (TPM) adapts the shared generative pathway to different fusion tasks among fusion stages. At the restoration stage, the degradation prompt router (DPR) retrieves input-adaptive degradation prompts to guide latent restoration. At the last stage, the application prompt bank (APB) injects application-specific semantic guidance throughout the decoding process. This functional hierarchy structurally separates the roles of task, degradation, and application conditions while coordinating them within a shared backbone, thereby enabling a unified model to support diverse fusion tasks, degradation conditions, and downstream applications.
The main contributions of this work are summarized as follows:
\begin{itemize}
    \item We propose HP-UniIF, a unified image fusion framework using diffusion priors to address the $\mathbf{M}^3$ challenge, simultaneously supporting multiple fusion tasks, robustness to diverse degradations, and adaptation to multiple downstream applications.

    \item We introduce a hierarchical prompt-learning comprising TPM, DPR, and APB. By organizing task-aware, degradation-aware, and application-specific prompts across hierarchical stages with a shared diffusion backbone for multiple fusion tasks, diverse degradation conditions, and multiple downstream applications.
    
    \item Extensive experiments show that HP-UniIF consistently improves both visual quality and downstream task performance across diverse fusion settings, degradation types, and application scenarios.
\end{itemize}
\section{Related Work}

\subsection{General Image Fusion Across Fusion Tasks}
Unified image fusion seeks to handle multimodal, multi-exposure, and multi-focus fusion within a single framework. Existing methods mainly differ in how cross-task knowledge is shared. IFCNN~\cite{zhang2020ifcnn} adopted a shared encoder-decoder with task-specific fusion rules, while U2Fusion~\cite{xu2020u2fusion} used unsupervised learning to preserve complementary source information. Subsequent methods improved shared representations through feature decomposition~\cite{zhang2021sdnet}, cross-domain interaction~\cite{ma2022swinfusion}, and reusable memory~\cite{cheng2023mufusion}. CCF~\cite{cao2024conditional} further introduced conditional diffusion for fusion-task control. However, these approaches primarily focus on task unification, with limited support for degradation-aware restoration, downstream adaptation, and input-adaptive specialization across representation levels.

% General image fusion aims to develop a unified framework capable of handling multiple fusion scenarios, including multi-modal, multi-exposure, and multi-focus fusion. Existing unified fusion methods mainly differ in how cross-task knowledge is shared. IFCNN~\cite{zhang2020ifcnn} employed a common encoder-decoder architecture together with task-dependent fusion rules, whereas U2Fusion~\cite{xu2020u2fusion} introduced an unsupervised objective to adaptively preserve complementary source information. Later methods improve shared representations through lightweight feature decomposition~\cite{zhang2021sdnet}, long-range cross-domain interaction~\cite{ma2022swinfusion}, and reusable memory representations~\cite{cheng2023mufusion}. The diffusion-based method  CCF~\cite{cao2024conditional} demonstrates the effectiveness of conditional diffusion for unified fusion, but it mainly focuses on fusion-task control and does not jointly coordinate degradation-aware restoration and downstream application adaptation. Nevertheless, existing approaches commonly accommodate different fusion tasks, leaving fine-grained, input-adaptive task specialization across representation levels insufficiently explored.
% Despite their effectiveness, these approaches largely rely on discriminative feature extraction and hand-crafted reconstruction objectives, limiting their ability to exploit the rich structural and semantic knowledge encoded in large-scale pretrained generative models.

\subsection{Image Fusion in Complex Environments}
Fusion models trained on clean images are vulnerable to diverse degradation conditions whose artifacts may propagate into fused outputs. Early methods addressed specific conditions: PIAFusion~\cite{tang2022piafusion} and DIVFusion~\cite{tang2023divfusion} targeted illumination degradation in infrared--visible fusion, while Deno-IF~\cite{xu2026deno} considered noisy observations. Later studies pursued unified degradation-aware fusion. Text-IF~\cite{yi2024text} used text guidance to control multiple corruptions, and DRMF~\cite{tang2024drmf} exploited composable diffusion priors for joint restoration and fusion. MMAIF~\cite{cao2025mmaif} further unified multiple fusion tasks and degradations through linguistic conditions, whereas URFusion~\cite{xu2025urfusion} learned degradation-invariant content and high-quality appearance. However, these methods remain primarily restoration-oriented, emphasizing perceptual fidelity rather than downstream-discriminative information.

% Fusion models trained predominantly on clean images are often vulnerable to real-world degradations, including poor illumination, noise, blur, and adverse weather, which can be directly inherited by the fused output. Early studies primarily addressed individual degradation types through scenario-specific designs. PIAFusion~\cite{tang2022piafusion} and DIVFusion~\cite{tang2023divfusion} incorporated illumination-aware fusion and joint low-light enhancement into infrared--visible fusion, respectively, while Deno-IF~\cite{xu2026deno} focused on unsupervised fusion under noisy observations. Subsequent efforts shifted toward unified degradation-aware fusion. Text-IF~\cite{yi2024text} introduced semantic text guidance to enable controllable handling of multiple corruptions, whereas DRMF~\cite{tang2024drmf} leveraged composable diffusion priors for joint restoration and multimodal fusion. Although MMAIF~\cite{cao2025mmaif} jointly considers multiple fusion tasks and degradation types, it relies on linguistic cues to degenerate semantics, while URFusion~\cite{xu2025urfusion} learned degradation-invariant content and high-quality appearance representations in an unsupervised manner. Existing methods remain predominantly restoration-oriented, emphasizing degradation removal and perceptual fidelity without explicitly preserving reliable cues for downstream perception.

\begin{figure*}[t]
    \centering
    \includegraphics[width=0.8\linewidth]{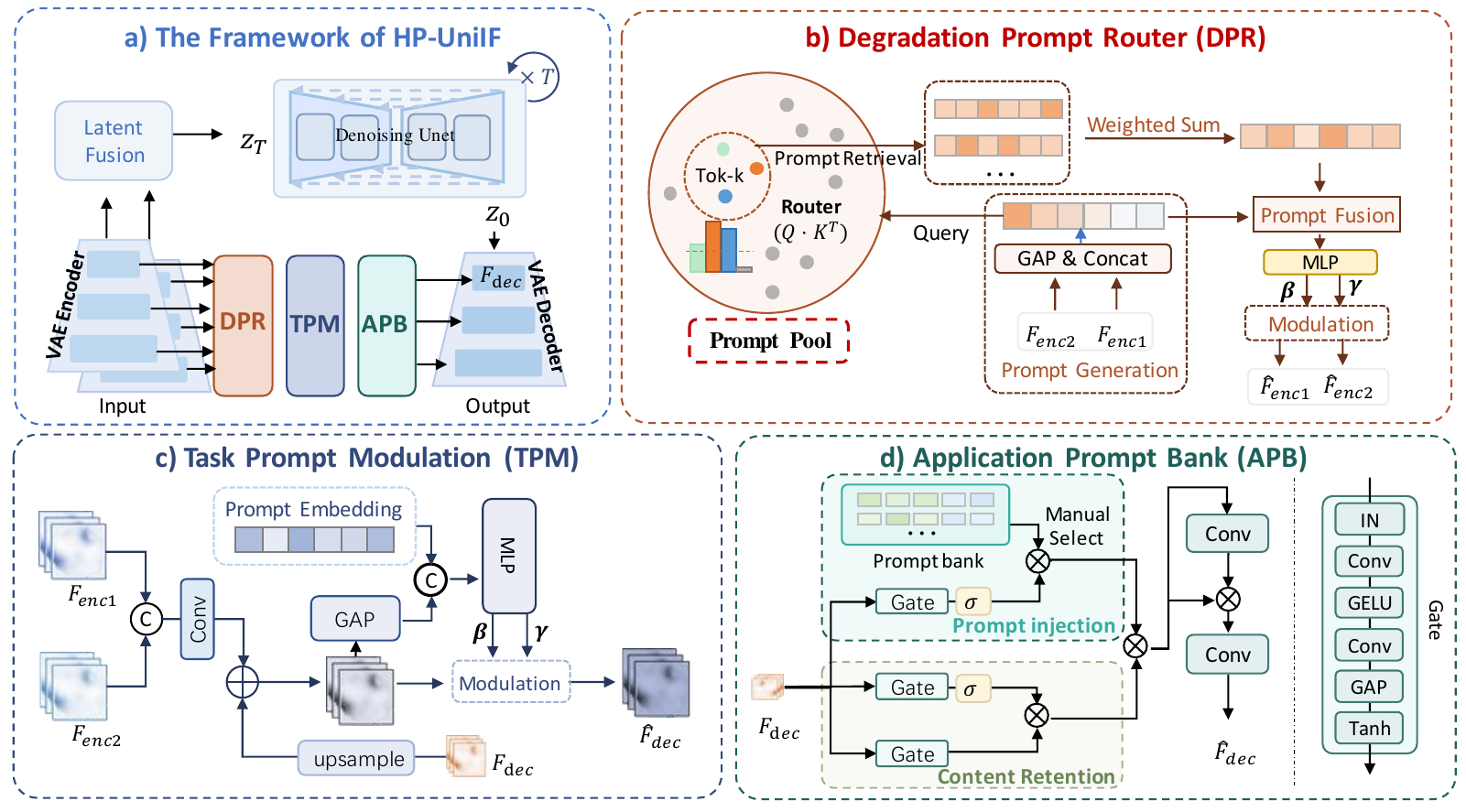}
    \caption{
    Overview of HP-UniIF. TPM progressively refines the fusion-task prompt to adapt the shared fusion pathway; DPR retrieves input-dependent degradation prompts for multi-scale restoration; and APB injects the specified downstream-application condition while retaining source evidence.
    }
    \label{fig:framework}
\end{figure*}

\subsection{Image Fusion for Downstream Applications}

Beyond perceptual quality, fused images increasingly support downstream perception. Early methods incorporated task supervision into fusion learning: TarDAL~\cite{liu2022target} jointly optimized fusion and object detection, while SeAFusion~\cite{tang2022image} introduced semantic supervision to preserve segmentation-relevant information. Later approaches strengthened task-oriented fusion through meta-learning~\cite{liu2024task}, high-order semantic interaction~\cite{zhou2024probing}, and direct downstream feedback~\cite{bai2025task}. SAGE~\cite{wu2025every} further distilled semantic priors from SAM to improve fusion quality and application adaptability. Nevertheless, most methods are optimized for a predefined application or tightly coupled with a fixed task network, making it difficult to flexibly reuse a shared fusion model across multiple downstream applications.

\section{Method}
\label{sec:method}

\subsection{Framework Overview}
\label{sec:method_overview}

Given two source images $\mathbf{x}^{1}$ and $\mathbf{x}^{2}$, HP-UniIF produces a fused image conditioned on a specified fusion task $t\in\mathcal{T}$ and an optional downstream application
$a\in\mathcal{A}\cup\{\varnothing\}$:
\begin{equation}
    \mathbf{y} = \mathcal{G}_{\Theta}(\mathbf{x}^{1},\mathbf{x}^{2};t, a).
    \label{eq:problem_formulation}
\end{equation}
The task identity $t$ and degradation $d$ are estimated adaptively from the input representations, while $a$ is provided only when application-oriented adaptation is required.

As illustrated in Fig.~\ref{fig:framework}, HP-UniIF is built on a pretrained latent diffusion model consisting of a variational autoencoder (VAE) and a denoising U-Net. The VAE encoder extracts a latent code and hierarchical skip features from each source image, and the two latent codes are fused to initialize the diffusion process:
\begin{equation}
    \mathbf{z}^{i}, \{\mathbf{s}^{i}_{l}\}_{l=1}^{L}=\mathcal{E}(\mathbf{x}^{i}),
    \qquad
    \mathbf{z}_{T}=\mathcal{F}_{z}([\mathbf{z}^{1};\mathbf{z}^{2}]).
    \label{eq:latent_encoding_fusion}
\end{equation}
Here, $[\cdot;\cdot]$ denotes channel-wise concatenation. Starting from$\mathbf{z}_{T}$, the denoising U-Net performs $T$ refinement steps to obtain $\mathbf{z}_{0}$, which is subsequently reconstructed by the VAE decoder in a coarse-to-fine manner.

At each decoder level, the Degradation Prompt Router (DPR) first restores the two source skip features before cross-source fusion. Task Prompt Modulation (TPM) then combines the restored features with the incoming decoder representation and performs task-conditioned modulation. When an application identity is provided, the Application Prompt Bank (APB) further adapts the representation before it enters the current decoder block.

\subsection{Task Prompt Modulation}
\label{sec:tpm}
Because different fusion tasks require distinct information-selection strategies, TPM employs task-aware prompts to progressively adapt the shared latent and decoder representations to the percept fusion task. For each task $t$, an initial prompt $\mathbf{p}^{T}_{0}$ is selected from a learnable task prompt bank. At decoder level $l$, the source features restored by DPR are fused and combined with the incoming decoder representation:
\begin{equation}
    \mathbf{u}^{T}_{l} = \mathcal{C}^{f}_{l}([
            \widetilde{\mathbf{s}}^{1}_{l};
            \widetilde{\mathbf{s}}^{2}_{l}
        ]
    )+\mathcal{U}^{T}_{l}
    (
        \mathbf{h}_{l+1}
    ),
    \label{eq:tpm_feature_aggregation}
\end{equation}
where $\mathcal{C}^{f}_{l}$ performs cross-source fusion and
$\mathcal{U}^{T}_{l}$ aligns the incoming decoder feature to the current resolution.

Instead of using a fixed task embedding throughout decoding, TPM progressively refines the prompt according to the current feature state:
\begin{equation}
    \mathbf{p}^{T}_{l+1} = \mathbf{p}^{T}_{l} +
    R^{T}_{l}
    ([ \mathbf{p}^{T}_{l} +\mathbf{e}^{T}_{l};\Phi^{T}_{l}(\operatorname{GAP}(\mathbf{u}^{T}_{l}))]),
    \label{eq:tpm_prompt_refinement}
\end{equation}
where $\mathbf{e}^{T}_{l}$ is a learnable stage token and $R^{T}_{l}$ denotes a lightweight refinement network. This stage-wise propagation allows the prompt to absorb feature-dependent information from coarse semantics to fine spatial details.

The updated prompt predicts channel-wise affine parameters and an injection gate. The task-adapted feature is computed as
\begin{align}
  {\gamma}^{T}_{l},\boldsymbol{\beta}^{T}_{l},\mathbf{g}^{T}_{l} =\operatorname{MLP}_{l}(\mathbf{p}^{T}_{l+1}),\\
    \widehat{\mathbf{u}}^{T}_{l}=\mathbf{u}^{T}_{l} +
    \sigma(\mathbf{g}^{T}_{l})\odot\rho_l    [\sigma(\gamma^{T}_{l})\odot\mathbf{u}^{T}_{l}
     +\sigma(\beta^{T}_{l})
    ],
    \label{eq:tpm_modulation}
\end{align}
where $\rho_l$ is a stage-dependent residual scale and $\sigma(\cdot)$ denote an active operation. The identity pathway, bounded affine parameters, and sigmoid gate jointly prevent excessive task-specific perturbations. The same feature-guided modulation is applied at the latent, decoder, and output stages using stage-specific parameters.
\subsection{Degradation Prompt Router}
\label{sec:dpr}
Degradations may affect the two source modalities differently and corrupt skip features at different spatial scales. Directly fusing such features can propagate unreliable evidence into the decoder. DPR therefore restores the two source skip features before cross-source fusion by combining reusable degradation prompts with an input-adaptive condition.

At decoder level $l$, DPR constructs a degradation descriptor from the two source features, their discrepancy, and the incoming decoder context:
\begin{equation}
    \mathbf{a}^{d}_{l}=\operatorname{GAP}
    (\mathcal{C}^{d}_{l} ([\mathbf{s}^{1}_{l}; \mathbf{s}^{2}_{l};
|\mathbf{s}^{1}_{l}-\mathbf{s}^{2}_{l}|;
\mathcal{U}^{d}_{l}(\mathbf{h}_{l+1})])).
    \label{eq:dpr_descriptor}
\end{equation}
The source discrepancy provides a direct cue for modality-dependent corruption, while the incoming decoder representation supplies coarse-level context. DPR maintains a level-specific pool of learnable key--value prompt pairs. A normalized query derived from $\mathbf{a}^{d}_{l}$ retrieves the $K$ most relevant entries, which are aggregated as
\begin{align}
    \mathbf{p}^{r}_{l}=\sum_{j\in\mathcal{I}_{l}} \frac{\exp(r^{d}_{l,j}/\tau_d)
    }{\sum_{m\in\mathcal{I}_{l}}\exp(r^{d}_{l,m}/\tau_d)}\mathbf{p}^{d}_{l,j}, \\
    \mathcal{I}_{l}=\operatorname{TopK}(\{r^{d}_{l,j}\}_{j=1}^{N_p}, K ),
    \label{eq:dpr_retrieval}
\end{align}
where $r^{d}_{l,j}$ is the cosine similarity between the query and the $j$-th prompt key, and $\tau_d$ is a temperature parameter.

The retrieved prompt is combined with an input-dependent prompt generated from $\mathbf{a}^{d}_{l}$ to form the degradation condition $\mathbf{c}^{d}_{l}$. Conditioned on this representation, a lightweight restoration block predicts a source-specific correction and its injection gate:
\begin{equation}
    \widetilde{\mathbf{s}}^{i}_{l}=\mathbf{s}^{i}_{l}
    +\sigma(\mathbf{g}^{d,i}_{l})
    \odot
    \Delta\mathbf{s}^{d,i}_{l},
   \Delta\mathbf{s}^{d,i}_{l},
        \mathbf{g}^{d,i}_{l}= \mathcal{R}^{d,i}_{l} ( \mathbf{s}^{i}_{l}, \mathbf{c}^{d}_{l}
    ).
    \label{eq:dpr_restoration}
\end{equation}
The residual pathway preserves reliable source evidence, while the learned gate controls the amount of degradation-aware correction. The restored features are subsequently fused and modulated by TPM.
\subsection{Application Prompt Bank}
\label{sec:apb}
Different downstream applications may favor different evidence in the fused representation. APB therefore introduces application-specific adaptation while keeping the shared fusion and restoration paths unchanged. For a specified application $a$, the corresponding prompt is selected from a learnable application prompt bank and propagated through the decoder. At level $l$, APB jointly processes the incoming decoder feature, the task-modulated skip representation, and the current application condition:
\begin{equation}
 \widetilde{\mathbf{h}}^{a}_{l},\mathbf{c}^{a}_{l+1} =\mathcal{A}_{l}
    (\mathbf{h}_{l+1},\widehat{\mathbf{u}}^{T}_{l},\mathbf{c}^{a}_{l}),
    \qquad
    \mathbf{c}^{a}_{0}=\mathbf{E}[a].
    \label{eq:apb_adaptation}
\end{equation}
The adapted feature $\hat{\mathbf{h}}_{l}^a$ is subsequently passed to the $l$-th decoder block.
As shown in Fig.~\ref{fig:framework}(d), $\mathcal{A}_{l}$ contains two complementary gated branches. The prompt-injection branch introduces application-relevant priors, while the content-retention branch preserves fusion evidence and selectively applies a lightweight content adapter. Their outputs are combined to update the decoder feature, and the application condition is propagated to the next stage.

When no application is specified, APB is bypassed and the task-modulated representation is directly passed to the decoder. In this work, APB is instantiated for object detection and semantic segmentation. New applications can be introduced by learning additional prompt entries and lightweight adapters while keeping the shared fusion backbone frozen.
\subsection{Training Strategy}
\label{sec:training}

HP-UniIF is trained in three stages to decouple fusion learning, degradation restoration, and application adaptation.

\noindent\textbf{S1: Multi-Task Fusion Learning.}
The latent and skip-fusion layers, decoder, and TPM are jointly trained on clean source pairs using the task-specific fusion objective $\mathcal{L}_{1}= \mathcal{L}^{t}_{\mathrm{fus}} $

\noindent\textbf{S2: Degradation-Aware Restoration.}
The Stage-1 fusion path is frozen, and DPR is optimized on paired clean and degraded samples:
\begin{equation}
    \mathcal{L}_{2} =\mathcal{L}_{1}+\sum_{l=0}^L \lambda_{l}(f^{clear}_i-f_l^{restore})
    \label{eq:stage2}
\end{equation}

where $L$ denotes the number of layers in the encoder, and $\lambda_i$ represents the scaling weight for the $i$th layer.

\noindent\textbf{S3: Application-specific Adaptation.}
The fusion and restoration pathways are frozen, while APB and the application-specific components are optimized using
$\mathcal{L}_{3}=\mathcal{L}_1+\lambda_{\mathrm{app}}\mathcal{L}_{\mathrm{app}}
$
where $\mathcal{L}{\mathrm{app}}$ denotes the application-specific objective computed from the fused output and its corresponding ground-truth annotation. Incorporating a new application into the prompt bank requires only the addition of an application-specific prompt, introducing a small number of trainable parameters that are optimized using the corresponding training data and objective. During inference, the appropriate application prompt is manually specified according to the target real-world scenario.
 The detailed losses are provided in Suppl.
\begin{figure*}[t]
\centering
\includegraphics[width=1\textwidth]{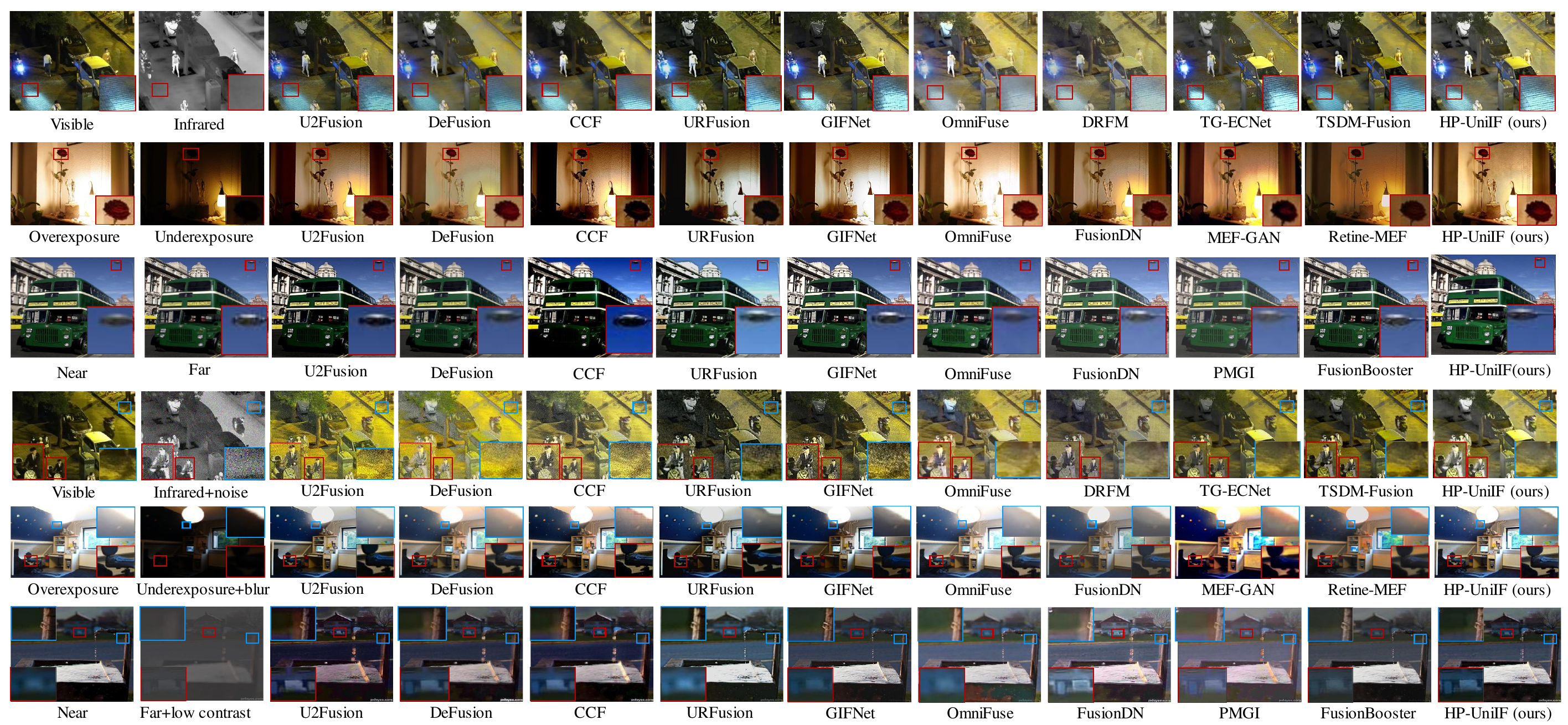} % Reduce the figure size so that it is slightly narrower than the column.
\caption{Qualitative comparison of multi-task image fusion under clean (row 1-3) and degraded (row 4-6) conditions. From top to bottom, the three rows present representative results for VIF, MEF, and MFF tasks, respectively.}
\label{fig:comparison_all}
\end{figure*}
\section{Experiments}
\subsection{Experimental Setup}
\noindent\textbf{Datasets}. We conduct experiments in three image fusion scenarios: VIF, MFF, and MEF. For VIF, we evaluate our model on the LLVIP~\cite{jia2021llvip} dataset, following the training and testing setting of~\cite{zhu2024task}. For MEF, we use 589 image pairs from SCIE~\cite{cai2018learning} for training and 100 image pairs from MEFB~\cite{zhang2021benchmarking} for testing. For each sequence in SCIE, the most underexposed and overexposed images are selected as the two inputs. For MFF, the model is trained on RealMFF~\cite{zhang2020real} and MFI-WHU~\cite{zhang2021mff} and follow the test setting in MFIF~\cite{zhang2021deep}

\noindent\textbf{Implementation Details.}
We initialize the diffusion backbone using the pretrained SD-Turbo model~\cite{sauer2024adversarial} and randomly crop the input images to $512\times512$ during training. More details are provided in Suppl.

\noindent\textbf{Competing Methods.}
We compare our method with three categories of representative baselines. First, we consider general image fusion methods, including U2Fusion~\cite{xu2020u2fusion}, DeFusion~\cite{liang2022fusion}, and CCF~\cite{cao2024conditional}. Second, we include recent generalized and degradation-aware fusion methods, including GIFNet~\cite{cheng2025one}, URFusion~\cite{xu2025urfusion}, and OmniFuse~\cite{zhang2025omnifuse}. Third, we compare with task-specific fusion methods for each fusion scenario. For visible--infrared fusion (VIF), the task-specific baselines include DRMF~\cite{tang2024drmf}, TG-ECNet~\cite{sun2025task}, RPFNet~\cite{guan2025residual}, SAGE~\cite{wu2025every}, and TSDM-Fusion~\cite{xue2024novel}, among which DRMF, TG-ECNet, and TSDM-Fusion explicitly address fusion under degraded inputs. For multi-exposure fusion MEF, we evaluate the aforementioned general methods together with MEF-GAN~\cite{xu2020mef} and Retinex-MEF~\cite{bai2025retinex}. For MFF, we additionally compare with FusionDN~\cite{xu2020fusiondn}, PMGI~\cite{zhang2020rethinking}, and FusionBooster~\cite{cheng2023fusionbooster}. To ensure a fair comparison under degraded inputs, OneRestore~\cite{guo2024onerestore} is used as a preprocessing module for task-specific methods that do not inherently incorporate degradation restoration, thereby forming a restoration-then-fusion pipeline. 

\noindent\textbf{Evaluation Metrics.}
% We evaluate the fusion results both qualitatively and quantitatively. Qualitative evaluation is based on visual inspection, focusing primarily on the preservation of salient structures, texture details, contrast, and natural appearance in the fused images. Quantitative evaluation measures both the intrinsic quality of each fused image and its information consistency with the corresponding source images.
Considering the distinct characteristics of different fusion tasks, we adopt task-specific evaluation metrics. For VIF, we employ six metrics: entropy (EN), gradient-based similarity measurement ($Q^{AB/F}$), mutual information (MI), visual information fidelity for fusion (VIFF), the sum of correlations of differences (SCD), and standard deviation (SD). For MFF, we use peak signal-to-noise ratio (PSNR), ($Q^{AB/F}$), VIFF, normalized mutual information (NMI), feature mutual information (FMI), and weighted fusion quality index ($Q_w$). For MEF, we report PSNR, MI, VIFF, multi-exposure fusion structural similarity (MEF-SSIM), and natural image quality evaluator (NIQE).  For downstream evaluation, all fusion methods are assessed using the same detection (yolo V5~\cite{yolov5}) or segmentation (Deeplab V3+~\cite{chen2018encoder}).
% \noindent\textbf{Downstream Evaluation.} For downstream evaluation, all fusion methods are assessed using the same detection (yolo V5~\cite{yolov5}) or segmentation (Deeplab V3+~\cite{chen2018encoder}) architecture, data split, and optimization protocol. Further architectural and training details are provided in the supplementary material.

\begin{figure*}[t]
\centering
\includegraphics[width=\textwidth]{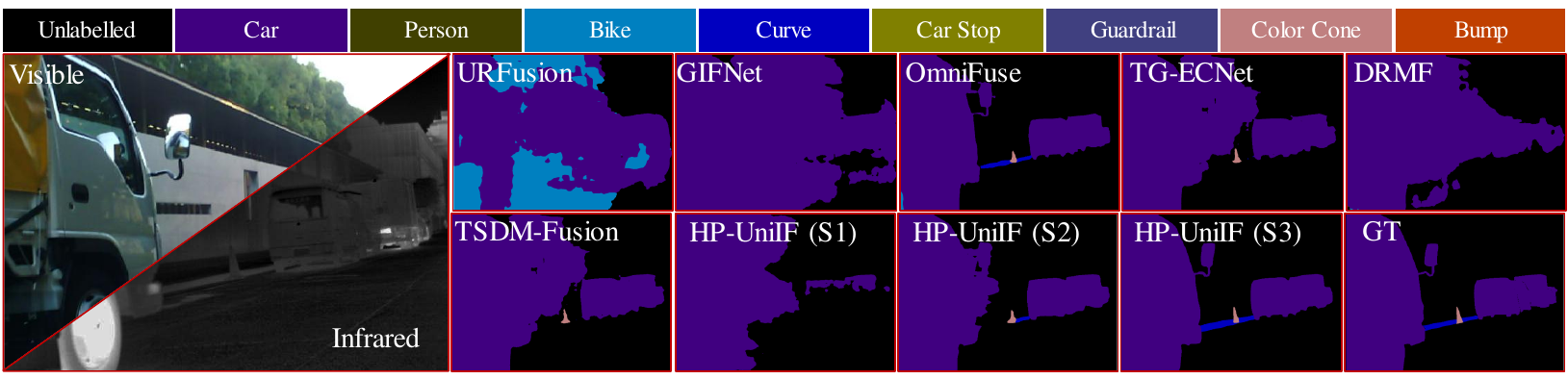}
\caption{
Qualitative comparison of downstream semantic segmentation on representative samples from the MSRS dataset. The segmentation predictions obtained from representative fusion methods and the three training stages of HP-UniIF, with the ground-truth segmentation shown in the final column.
}
\label{fig:segmentation}
\end{figure*}

\begin{figure*}[t]
\centering
\includegraphics[width=1\textwidth]{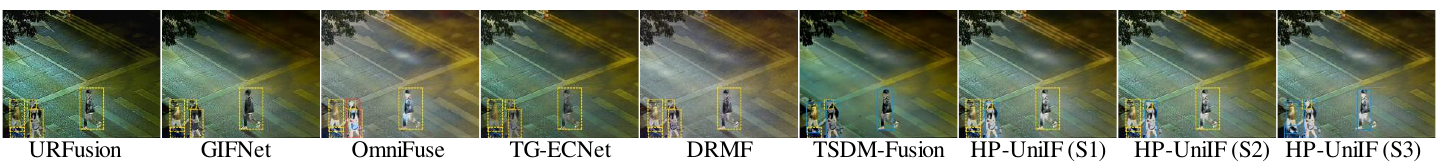} % Reduce the figure size so that it is slightly narrower than the column.
\caption{Qualitative comparison of downstream object detection on representative samples from the LLVIP dataset. The detection predictions obtained from representative fusion methods and the three training stages of HP-UniIF. Blue solid boxes indicate true positives, yellow dashed boxes indicate false negatives, and red dashed boxes indicate false positives.}
\label{fig:detection}
\end{figure*}
\subsection{Multi-Task Fusion without Degradation}
We conduct comprehensive qualitative and quantitative evaluations to compare HP-UniIF with recent methods across multiple fusion tasks under clean conditions.
Fig.~\ref{fig:comparison_all} (rows 1--3) presents qualitative
comparisons under clean input conditions. For VIF, several competing methods either weaken the thermal saliency of pedestrians or introduce over-smoothed road textures and unnatural color shifts. In contrast, HP-UniIF simultaneously preserves salient infrared targets, visible-spectrum appearance, and fine pavement structures. For MEF, HP-UniIF recovers informative content from the underexposed regions while avoiding severe saturation around the lamp and window. For MFF, it produces consistently focused structures in both foreground and background regions, with clearer vehicle contours and local details.
These results indicate that HP-UniIF can accommodate distinct
fusion objectives without requiring task-specific architectures.
The quantitative results are provided in the Suppl. and exhibit a consistent performance advantage.

\subsection{Multi-Task Fusion under Degraded Conditions}
\begin{table}[t]
\centering
\caption{Quantitative results of the VIF task under degraded conditions on the LLVIP dataset. The best and second-best results are highlighted in \textbf{bold} and \underline{underlined}, respectively.}
\label{tab:vif_degraded}
\small
\setlength{\tabcolsep}{2.8pt}
\resizebox{\columnwidth}{!}{
\begin{tabular}{lcccccc}
\toprule
Method & EN$\uparrow$ & $Q_{\mathrm{AB/F}}\uparrow$ & MI$\uparrow$ & VIFF$\uparrow$ & SCD$\uparrow$& SD$\uparrow$ \\
\midrule
U2Fusion (TPAMI'20)
& 7.285& 0.345& 1.191& 0.428& 1.215& 47.034 \\
DeFusion (ECCV'22)
& \underline{7.363}& 0.297& 1.337& 0.442& 1.048& \underline{48.887} \\
CCF (NeurIPS'24)
& 7.278& 0.326& 1.230& 0.422& 1.187& 48.150 \\
URFusion (TIP'25)
& 7.174 & 0.384& 1.312& 0.430& 1.214 & 48.359 \\
GIFNet (CVPR'25)
& 6.867& 0.306& 1.303& 0.375& 1.188& 42.090 \\
OmniFuse (TPAMI'25)
& 7.147& 0.377& 1.478& 0.424& 1.050& 44.618 \\
\midrule
DRMF (ACM MM'24)
& 7.252& 0.342& \underline{1.845}& 0.459& 1.134& 46.210 \\
TG-ECNet (ICML'25)
& 7.212& \underline{0.470}& 1.641& \underline{0.555}& \underline{1.350}& 47.864\\
% RPFNet (ACM MM'25)
	% & 6.418 & 0.320 & 1.285 &0.501 &1.233 &48.111\\
% SAGE (CVPR'25)
 % & 6.540 & 0.281 & 1.293 & 0.580 & 1.228&\textbf{62.770}\\
TSDM-Fusion (TIM'26)
 & 6.971& 0.357& 1.424& 0.457& 1.192&39.652 \\

\midrule

HP-UniIF & \textbf{7.381} & \textbf{0.487} & \textbf{1.852} & \textbf{0.578} & \textbf{1.619}& \textbf{49.919} \\
\bottomrule
\end{tabular}
}
\end{table}
\begin{table}[t]
\centering
\caption{
Quantitative comparison on the MEF task under degraded conditions. The best and second-best results are highlighted in  \textbf{bold} and \underline{underline},
respectively.
}
\label{tab:mef_degraded}
\small
\setlength{\tabcolsep}{3.2pt}
\resizebox{\columnwidth}{!}{
\begin{tabular}{lccccc}
\toprule
Method
& PSNR$\uparrow$
& MI$\uparrow$
& VIFF$\uparrow$
& MEF-SSIM$\uparrow$
& NIQE$\downarrow$ \\
\midrule

% \multicolumn{6}{l}{\textit{General Image Fusion Methods}} \\
U2Fusion (TPAMI'20)
& 21.918& 4.603& \underline{1.051}& \underline{0.841}& 4.689 \\
DeFusion (ECCV'22)
& \underline{23.879} & 5.224& 0.924& 0.830& 4.602 \\
CCF (NeurIPS'24)
& 22.479 & \underline{5.262} & 0.957 & 0.809 & 5.016 \\
URFusion (TIP'25)
& 22.530& 5.121& 0.832& 0.813& 4.975 \\
GIFNet (CVPR'25)
& 22.426& 5.116& 0.952& 0.810& 4.654 \\
OmniFuse (TPAMI'25)
& \textit{24.364}& \textit{5.279}& 0.883& 0.818& 5.766 \\
\midrule
% \multicolumn{6}{l}{\textit{MEF-specific Methods}} \\
FusionDN (AAAI'20)
& 21.368& 5.692& 1.032& 0.830& 4.528 \\
MEF-GAN (TIP'20)
& 22.136 & 4.954& 0.621& 0.708& \underline{4.431} \\
Retinex-MEF (CVPR'25)
& 21.763& 5.257& 0.936& 0.809& 5.100 \\
\midrule
% \multicolumn{6}{l}{\textit{Ours}} \\
HP-UniIF (ours)
& \textbf{24.633}& \textbf{5.331}& \textbf{1.190}& \textbf{0.920}& \textbf{3.791} \\

\bottomrule
\end{tabular}
}
\end{table}

\begin{table*}[t]
\centering
\caption{
Quantitative comparison of downstream object detection and semantic segmentation. Per-class segmentation results are measured by IoU. The best and second-best results are highlighted in \textbf{bold} and \underline{underlined}, respectively.
}
\label{tab:downstream}
\setlength{\tabcolsep}{3.0pt}
\resizebox{\textwidth}{!}{
\begin{tabular}{lcc|cccccccccc}
\toprule
\multirow{2}{*}{Method}
& \multicolumn{2}{c|}{Detection}
& \multicolumn{10}{c}{Segmentation} \\
\cmidrule(lr){2-3}
\cmidrule(lr){4-13}
& mAP@.5& mAP@.5:.95& Car& Person& Bike& Curve& Car Stop& Guardrail& Color Cone& Bump& mIoU& mAcc \\
\midrule

URFusion (TIP'25)
& 0.719&0.485& 84.717& 68.380& 64.069& 50.807& 67.359& 64.181& 53.240& 70.303& 69.009& 74.846 \\
GIFNet (CVPR'25)
& 0.342& 0.207& 88.022& 70.329& 70.267& 54.733& 70.361& 70.961& 64.749& 60.774& 72.050& 78.265 \\
OmniFuse (TPAMI'25)
&0.711&0.445& 85.694& 64.197& 61.780& 47.668& 67.164& 70.326& 58.671& 62.097& 68.393& 73.576 \\
TG-ECNet (ICML'25)
& 0.605& 0.415& 87.518& 70.291& 67.410& 50.887& 67.835& 69.326& 61.045& 67.987& 71.167& 76.817 \\
DRMF (ACM MM'24)
& 0.547& 0.358& 79.553& 66.708& 60.099& 26.414& 49.107& 45.675& 54.498& 17.089& 55.162& 58.714 \\
TSDM-Fusion (TIM'26)
&\underline{0.859}&0.522& 87.378& \underline{71.724}& 68.458& 53.384& 67.553
& 54.645& 64.070& 59.157& 69.395& 75.182 \\

\midrule
HP-UniIF (S1)
& 0.817& \underline{0.583}& 89.547& 71.471& \textbf{71.022}& 62.231& 70.141
& \underline{72.903}& 63.880& \underline{75.316}& \underline{74.994}& \underline{81.932} \\

HP-UniIF (S2)	

& 0.829& 0.521& \underline{89.673}& 71.646
& \underline{70.840}& \textbf{62.628}& \textbf{70.763}& 70.728
& \underline{64.814}& 72.528& 74.673& 81.810 \\

HP-UniIF (S3)
& \textbf{0.864}& \textbf{0.618}& \textbf{89.730}& \textbf{72.366}
& 69.417& \underline{62.534}& \underline{70.457}& \textbf{73.287}& \textbf{64.900}& \textbf{78.928}& \textbf{75.564}& \textbf{82.970} \\
\bottomrule
\end{tabular}
}
\end{table*}
\begin{table}[t]
\centering
\caption{
Quantitative comparison on the MFF task under degraded conditions.
The best and second-best results are highlighted in \textbf{bold} and \underline{underline},
respectively.
}
\label{tab:mff_degraded}
\small
\setlength{\tabcolsep}{3.2pt}
\resizebox{\columnwidth}{!}{
\begin{tabular}{lcccccc}
\toprule
Method
& PSNR$\uparrow$
& $Q_{\mathrm{AB/F}}\uparrow$
& VIFF$\uparrow$
& NMI$\uparrow$
& FMI$\uparrow$
& $Q_W\uparrow$ \\
\midrule

U2Fusion (TPAMI'20)
& 45.573& \textit{0.507}& \textit{0.884}& 0.849& 0.710& \textit{0.729}\\
DeFusion (ECCV'22)
& \textit{48.723}& 0.408& 0.868& \textit{0.886}& 0.722& 0.560 \\
CCF (NeurIPS'24)
& 44.400 & 0.390& 0.862& 0.849& \textit{0.735}& 0.547 \\
URFusion (TIP'25)
& 31.399& 0.395& 0.722& 0.843& 0.720& 0.602 \\
GIFNet (CVPR'25)
& 41.099& 0.432& 0.785& 0.858& 0.729& 0.639 \\
OmniFuse (TPAMI'25)
& 40.315& 0.341& 0.705& 0.760& 0.720& 0.443 \\
\midrule

FusionDN (AAAI'20)
& 40.003 & 0.470 & 0.859& 0.850& 0.705&0.686  \\
PMGI (AAAI'20)
& 35.616& 0.377& 0.544& 0.755& 0.688& 0.544 \\
FusionBooster (TPAMI'25)
& 42.569& 0.479& 0.842& 0.872& 0.713& 0.685 \\
\midrule

HP-UniIF (ours) & \textbf{52.689}& \textbf{0.586}& \textbf{1.021}& \textbf{0.906}& \textbf{0.778}& \textbf{0.790} \\

\bottomrule
\end{tabular}
}
\end{table}

Fig.~\ref{fig:comparison_all} (rows 4--6) presents qualitative
comparisons on the three fusion tasks with degraded source inputs. For VIF, several competing methods either propagate the severe noise contained in the source images or introduce noticeable color distortion and structural blurring. In contrast, HP-UniIF suppresses noise more effectively while retaining clear pedestrian silhouettes, vehicle boundaries, and background textures. For MEF, existing methods tend to leave dark regions insufficiently exposed or produce saturation artifacts around high-intensity areas such as the lamp and window. HP-UniIF achieves a more balanced exposure, with clearer structures in both bright and dark regions. For MFF, some methods retain residual defocus blur or introduce inconsistent brightness and color shifts across focused regions. HP-UniIF produces sharper foreground and background structures, particularly around the pole, building contours, road markings, and other fine boundaries. These observations demonstrate that HP-UniIF can adapt to different task-specific fusion objectives while consistently mitigating heterogeneous degradations.

The quantitative results in Table~\ref{tab:vif_degraded}, Table~\ref{tab:mef_degraded} and Table~\ref{tab:mff_degraded} further validate the effectiveness of HP-UniIF. It achieves the best performance on all 17 metrics across the three fusion tasks. 
% On degraded VIF, HP-UniIF obtains an EN of 7.381, $Q^{AB/F}$ of 0.487, MI of 1.852, VIFF of 0.578, SCD of 1.619, and SD of 49.919. In particular, it improves SCD and VIFF over the corresponding second-best results by 19.9\% and 4.1\%, respectively, indicating more effective preservation of complementary source information and perceptually meaningful structures. On degraded MFF, HP-UniIF outperforms the strongest competing results by 3.966 dB in PSNR and achieves relative improvements of 15.6\%, 15.5\%, and 8.4\% in $Q^{AB/F}$, VIFF, and $Q_W$, respectively. These improvements reflect its ability to recover consistently focused structures while reducing degradation-induced artifacts. On degraded MEF, HP-UniIF improves VIFF and MEF-SSIM by 13.2\% and 9.4\%, respectively, while reducing NIQE from 4.431 to 3.791, corresponding to a 14.4\% improvement. 
The simultaneous gains in fidelity, structural preservation, information retention, and perceptual quality show that the improvements do not arise merely from increased contrast or residual noise. Overall, the qualitative and quantitative results consistently demonstrate the robustness and task generalization ability of HP-UniIF under diverse degraded conditions.

% \begin{figure}[t]
% \centering
% \includegraphics[width=\columnwidth]{images/comprison2.png} % Reduce the figure size so that it is slightly narrower than the column.
% \caption{Qualitative comparison of image fusion in the LLVIP dataset.}
% \label{fig:comprison_clean}
% \end{figure}

\subsection{Comparison on Downstream Applications}
We select several representative fusion methods for downstream evaluation in visible-infrared object detection and semantic segmentation scenarios. Table~\ref{tab:downstream} compares the downstream performance of different fusion methods on both tasks. HP-UniIF achieves the best overall results in object detection and semantic segmentation. In particular, its more pronounced improvement in mAP@.5:.95 indicates that HP-UniIF not only preserves discriminative target cues but also enables more accurate object localization. For semantic segmentation, HP-UniIF obtains the highest mIoU and mAcc and achieves the best class-wise IoU on five of the eight categories, covering both large semantic regions and small foreground objects. APB injects application-specific semantic guidance while largely preserving complementary fusion evidence, enabling the fused images to better support downstream applications. Fig.~\ref{fig:segmentation} and Fig.~\ref{fig:detection} provide qualitative support, with additional comparisons presented in the Suppl.

% Compared with competing methods and the earlier training stages, HP-UniIF produces fewer missed detections, more complete semantic regions, and cleaner boundaries for small objects. The consistent improvements from Stage 1 to Stage 3 demonstrate that application-aware adaptation effectively bridges perceptual fusion quality and downstream semantic utility.

\subsection{Ablation Studies}
We conduct component-level and stage-wise ablation studies to evaluate the contributions of the proposed hierarchical design. Component-level ablations are first performed on the LLVIP dataset using intrinsic fusion metrics. To provide a more comprehensive assessment, we further evaluate the effects of individual components on downstream object detection and semantic segmentation.
\begin{table}[t]
\centering
\caption{
Ablation studies of HP-UniIF on the LLVIP dataset. The best and second-best results are highlighted in \textbf{bold} and \underline{underlined}, respectively. All metrics are higher-is-better.
}
\label{tab:component_ablation}
\small
\setlength{\tabcolsep}{1.7pt}
\begin{tabular}{cccccccccc}
\toprule
DIFF.& TPM& DPR& APB& EN& $Q^{AB/F}$& MI& VIFF& SCD& SD \\
\midrule
--& --& --& --& \textbf{7.484}& 0.501& 1.611& 0.477& 1.563& \underline{52.48} \\
$\checkmark$& --& --& --& 7.382& 0.592& 1.970& 0.662& 1.572& 49.79 \\
$\checkmark$& $\checkmark$& --& --& 7.421& 0.591& \textbf{2.025}& \underline{0.670}& \underline{1.618}& 51.42 \\
$\checkmark$& --& $\checkmark$& --& 7.388& 0.410& 1.743& 0.525& 1.599& 49.70 \\
% $\checkmark$& --& --& $\checkmark$& 7.283& 0.083& 0.419& 0.050& 0.037& \textbf{62.54} \\

$\checkmark$& $\checkmark$& $\checkmark$& --& \underline{7.467}& \textbf{0.605}& 1.965
& \textbf{0.693}& \textbf{1.619}& \underline{52.48} \\
$\checkmark$& $\checkmark$& $\checkmark$& $\checkmark$& 7.409& \underline{0.593}& \underline{2.006}& 0.667& 1.552& 50.58 \\

\bottomrule
\end{tabular}
\end{table}
As shown in Table~\ref{tab:component_ablation}, introducing the diffusion prior improves information preservation and perceptual fidelity. Compared with a vanilla U-Net trained from scratch without diffusion-based refinement, the pretrained diffusion backbone provides a stronger structural and generative prior for unified image fusion.
TPM further adapts this shared prior to different fusion objectives, whereas DPR alone provides limited benefits because degradation correction without task awareness may alter useful fusion evidence. When TPM and DPR are jointly employed, the model achieves the strongest overall fusion performance, confirming that task-aware fusion and degradation-aware restoration are complementary.
Adding APB causes only minor variations in intrinsic fusion quality but leads to clear improvements in downstream detection and segmentation. This indicates that application-specific semantics can be introduced without substantially disrupting the visual representation learned by the fusion pathway.

\section{Conclusion}
In this paper, we introduce HP-UniIF, a unified framework for image fusion using diffusion priors, jointly supporting heterogeneous fusion tasks, diverse degradations, and multiple downstream applications. HP-UniIF employs hierarchical prompt learning to organize task-aware fusion, degradation-aware restoration, and application-specific adaptation across complementary representation stages, thereby mitigating potential interference among heterogeneous objectives. Extensive experiments on VIF, MEF, and MFF demonstrate its effectiveness under both clean and degraded conditions, while downstream evaluations further validate its advantages for object detection and semantic segmentation. In the future, we will explore adaptation to unseen degradations and broader applications, as well as more lightweight prompt learning for efficient real-world deployment.

\bibliography{aaai2027}

\begin{thebibliography}{57}
\providecommand{\natexlab}[1]{#1}

\bibitem[{Bai et~al.(2025{\natexlab{a}})Bai, Zhang, Zhao, Deng, Cui, and Xu}]{bai2025retinex}
Bai, H.; Zhang, J.; Zhao, Z.; Deng, L.; Cui, Y.; and Xu, S. 2025{\natexlab{a}}.
\newblock Retinex-mef: Retinex-based glare effects aware unsupervised multi-exposure image fusion.
\newblock In \emph{Proceedings of the IEEE/CVF International Conference on Computer Vision}, 7251--7261.

\bibitem[{Bai et~al.(2025{\natexlab{b}})Bai, Zhang, Zhao, Wu, Deng, Cui, Feng, and Xu}]{bai2025task}
Bai, H.; Zhang, J.; Zhao, Z.; Wu, Y.; Deng, L.; Cui, Y.; Feng, T.; and Xu, S. 2025{\natexlab{b}}.
\newblock Task-driven image fusion with learnable fusion loss.
\newblock In \emph{Proceedings of the Computer Vision and Pattern Recognition Conference}, 7457--7468.

\bibitem[{Cai, Gu, and Zhang(2018)}]{cai2018learning}
Cai, J.; Gu, S.; and Zhang, L. 2018.
\newblock Learning a deep single image contrast enhancer from multi-exposure images.
\newblock \emph{IEEE transactions on image processing}, 27(4): 2049--2062.

\bibitem[{Cao et~al.(2024)Cao, Xu, Zhu, Wang, and Hu}]{cao2024conditional}
Cao, B.; Xu, X.; Zhu, P.; Wang, Q.; and Hu, Q. 2024.
\newblock Conditional controllable image fusion.
\newblock \emph{Advances in Neural Information Processing Systems}, 37: 120311--120335.

\bibitem[{Cao et~al.(2025)Cao, Zhong, Wang, and Deng}]{cao2025mmaif}
Cao, Z.; Zhong, Y.; Wang, Z.; and Deng, L.-J. 2025.
\newblock Mmaif: Multi-task and multi-degradation all-in-one for image fusion with language guidance.
\newblock In \emph{Proceedings of the IEEE/CVF International Conference on Computer Vision}, 11744--11754.

\bibitem[{Chen et~al.(2018)Chen, Zhu, Papandreou, Schroff, and Adam}]{chen2018encoder}
Chen, L.-C.; Zhu, Y.; Papandreou, G.; Schroff, F.; and Adam, H. 2018.
\newblock Encoder-decoder with atrous separable convolution for semantic image segmentation.
\newblock In \emph{Proceedings of the European conference on computer vision (ECCV)}, 801--818.

\bibitem[{Cheng et~al.(2025)Cheng, Xu, Feng, Wu, Tang, Li, Zhang, Atito, Awais, and Kittler}]{cheng2025one}
Cheng, C.; Xu, T.; Feng, Z.; Wu, X.; Tang, Z.; Li, H.; Zhang, Z.; Atito, S.; Awais, M.; and Kittler, J. 2025.
\newblock One model for all: Low-level task interaction is a key to task-agnostic image fusion.
\newblock In \emph{Proceedings of the IEEE/CVF Conference on Computer Vision and Pattern Recognition}, 28102--28112.

\bibitem[{Cheng, Xu, and Wu(2023)}]{cheng2023mufusion}
Cheng, C.; Xu, T.; and Wu, X.-J. 2023.
\newblock MUFusion: A general unsupervised image fusion network based on memory unit.
\newblock \emph{Information Fusion}, 92: 80--92.

\bibitem[{Cheng et~al.(2023)Cheng, Xu, Wu, Li, Li, and Kittler}]{cheng2023fusionbooster}
Cheng, C.; Xu, T.; Wu, X.-J.; Li, H.; Li, X.; and Kittler, J. 2023.
\newblock Fusionbooster: A unified image fusion boosting paradigm.
\newblock \emph{arXiv preprint arXiv:2305.05970}.

\bibitem[{Guan et~al.(2025)Guan, Wang, Qian, Liu, and Ma}]{guan2025residual}
Guan, Z.; Wang, X.; Qian, W.; Liu, P.; and Ma, R. 2025.
\newblock Residual prior-driven frequency-aware network for image fusion.
\newblock In \emph{Proceedings of the 33rd ACM International Conference on Multimedia}, 1082--1091.

\bibitem[{Guo et~al.(2024)Guo, Gao, Lu, Zhu, Liu, and He}]{guo2024onerestore}
Guo, Y.; Gao, Y.; Lu, Y.; Zhu, H.; Liu, R.~W.; and He, S. 2024.
\newblock Onerestore: A universal restoration framework for composite degradation.
\newblock In \emph{European conference on computer vision}, 255--272. Springer.

\bibitem[{James and Dasarathy(2014)}]{james2014medical}
James, A.~P.; and Dasarathy, B.~V. 2014.
\newblock Medical image fusion: A survey of the state of the art.
\newblock \emph{Information fusion}, 19: 4--19.

\bibitem[{Jia et~al.(2021)Jia, Zhu, Li, Tang, and Zhou}]{jia2021llvip}
Jia, X.; Zhu, C.; Li, M.; Tang, W.; and Zhou, W. 2021.
\newblock LLVIP: A visible-infrared paired dataset for low-light vision.
\newblock In \emph{Proceedings of the IEEE/CVF international conference on computer vision}, 3496--3504.

\bibitem[{Jocher(2020)}]{yolov5}
Jocher, G. 2020.
\newblock YOLOv5 by Ultralytics.
\newblock \mbox{doi}:\url{10.5281/zenodo.3908559}.

\bibitem[{Li and Wu(2018)}]{li2018densefuse}
Li, H.; and Wu, X.-J. 2018.
\newblock DenseFuse: A fusion approach to infrared and visible images.
\newblock \emph{IEEE transactions on image processing}, 28(5): 2614--2623.

\bibitem[{Li, Wu, and Kittler(2021)}]{li2021rfn}
Li, H.; Wu, X.-J.; and Kittler, J. 2021.
\newblock RFN-Nest: An end-to-end residual fusion network for infrared and visible images.
\newblock \emph{Information Fusion}, 73: 72--86.

\bibitem[{Liang et~al.(2022)Liang, Jiang, Liu, and Ma}]{liang2022fusion}
Liang, P.; Jiang, J.; Liu, X.; and Ma, J. 2022.
\newblock Fusion from decomposition: A self-supervised decomposition approach for image fusion.
\newblock In \emph{European conference on computer vision}, 719--735. Springer.

\bibitem[{Liu et~al.(2022)Liu, Fan, Huang, Wu, Liu, Zhong, and Luo}]{liu2022target}
Liu, J.; Fan, X.; Huang, Z.; Wu, G.; Liu, R.; Zhong, W.; and Luo, Z. 2022.
\newblock Target-aware dual adversarial learning and a multi-scenario multi-modality benchmark to fuse infrared and visible for object detection.
\newblock In \emph{Proceedings of the IEEE/CVF conference on computer vision and pattern recognition}, 5802--5811.

\bibitem[{Liu et~al.(2023)Liu, Liu, Wu, Ma, Liu, Zhong, Luo, and Fan}]{liu2023multi}
Liu, J.; Liu, Z.; Wu, G.; Ma, L.; Liu, R.; Zhong, W.; Luo, Z.; and Fan, X. 2023.
\newblock Multi-interactive feature learning and a full-time multi-modality benchmark for image fusion and segmentation.
\newblock In \emph{Proceedings of the IEEE/CVF international conference on computer vision}, 8115--8124.

\bibitem[{Liu et~al.(2024)Liu, Liu, Liu, Fan, and Luo}]{liu2024task}
Liu, R.; Liu, Z.; Liu, J.; Fan, X.; and Luo, Z. 2024.
\newblock A task-guided, implicitly-searched and meta-initialized deep model for image fusion.
\newblock \emph{IEEE Transactions on Pattern Analysis and Machine Intelligence}, 46(10): 6594--6609.

\bibitem[{Liu et~al.(2017)Liu, Chen, Peng, and Wang}]{liu2017multi}
Liu, Y.; Chen, X.; Peng, H.; and Wang, Z. 2017.
\newblock Multi-focus image fusion with a deep convolutional neural network.
\newblock \emph{Information Fusion}, 36: 191--207.

\bibitem[{Liu et~al.(2020)Liu, Wang, Cheng, Li, and Chen}]{liu2020multi}
Liu, Y.; Wang, L.; Cheng, J.; Li, C.; and Chen, X. 2020.
\newblock Multi-focus image fusion: A survey of the state of the art.
\newblock \emph{Information Fusion}, 64: 71--91.

\bibitem[{Ma, Ma, and Li(2019)}]{ma2019infrared}
Ma, J.; Ma, Y.; and Li, C. 2019.
\newblock Infrared and visible image fusion methods and applications: A survey.
\newblock \emph{Information fusion}, 45: 153--178.

\bibitem[{Ma et~al.(2022)Ma, Tang, Fan, Huang, Mei, and Ma}]{ma2022swinfusion}
Ma, J.; Tang, L.; Fan, F.; Huang, J.; Mei, X.; and Ma, Y. 2022.
\newblock SwinFusion: Cross-domain long-range learning for general image fusion via swin transformer.
\newblock \emph{IEEE/CAA Journal of Automatica Sinica}, 9(7): 1200--1217.

\bibitem[{Ma et~al.(2019)Ma, Yu, Liang, Li, and Jiang}]{ma2019fusiongan}
Ma, J.; Yu, W.; Liang, P.; Li, C.; and Jiang, J. 2019.
\newblock FusionGAN: A generative adversarial network for infrared and visible image fusion.
\newblock \emph{Information fusion}, 48: 11--26.

\bibitem[{Ma et~al.(2017)Ma, Li, Yong, Wang, Meng, and Zhang}]{ma2017robust}
Ma, K.; Li, H.; Yong, H.; Wang, Z.; Meng, D.; and Zhang, L. 2017.
\newblock Robust multi-exposure image fusion: a structural patch decomposition approach.
\newblock \emph{IEEE Transactions on Image Processing}, 26(5): 2519--2532.

\bibitem[{Ma et~al.(2023)Ma, Wang, Li, Yang, Li, Song, and Li}]{ma2023infrared}
Ma, W.; Wang, K.; Li, J.; Yang, S.~X.; Li, J.; Song, L.; and Li, Q. 2023.
\newblock Infrared and visible image fusion technology and application: A review.
\newblock \emph{Sensors}, 23(2): 599.

\bibitem[{Sauer et~al.(2024)Sauer, Lorenz, Blattmann, and Rombach}]{sauer2024adversarial}
Sauer, A.; Lorenz, D.; Blattmann, A.; and Rombach, R. 2024.
\newblock Adversarial diffusion distillation.
\newblock In \emph{European Conference on Computer Vision}, 87--103. Springer.

\bibitem[{Sun et~al.(2025)Sun, Li, Zhu, Hu, Ren, Xu, and Zhu}]{sun2025task}
Sun, Y.; Li, X.; Zhu, P.; Hu, Q.; Ren, D.; Xu, H.; and Zhu, X. 2025.
\newblock Task-gated multi-expert collaboration network for degraded multi-modal image fusion.
\newblock In \emph{International Conference on Machine Learning}, 57571--57586. PMLR.

\bibitem[{Sun et~al.(2023)Sun, Meng, Wang, Tang, Shen, and Wang}]{sun2023visible}
Sun, Y.; Meng, Y.; Wang, Q.; Tang, M.; Shen, T.; and Wang, Q. 2023.
\newblock Visible and infrared image fusion for object detection: a survey.
\newblock In \emph{International Conference on Image, Vision and Intelligent Systems}, 236--248. Springer.

\bibitem[{Tang et~al.(2024)Tang, Deng, Yi, Yan, Yuan, and Ma}]{tang2024drmf}
Tang, L.; Deng, Y.; Yi, X.; Yan, Q.; Yuan, Y.; and Ma, J. 2024.
\newblock DRMF: Degradation-robust multi-modal image fusion via composable diffusion prior.
\newblock In \emph{Proceedings of the 32nd ACM International Conference on Multimedia}, 8546--8555.

\bibitem[{Tang et~al.(2023)Tang, Xiang, Zhang, Gong, and Ma}]{tang2023divfusion}
Tang, L.; Xiang, X.; Zhang, H.; Gong, M.; and Ma, J. 2023.
\newblock DIVFusion: Darkness-free infrared and visible image fusion.
\newblock \emph{Information Fusion}, 91: 477--493.

\bibitem[{Tang, Yuan, and Ma(2022)}]{tang2022image}
Tang, L.; Yuan, J.; and Ma, J. 2022.
\newblock Image fusion in the loop of high-level vision tasks: A semantic-aware real-time infrared and visible image fusion network.
\newblock \emph{Information Fusion}, 82: 28--42.

\bibitem[{Tang et~al.(2022)Tang, Yuan, Zhang, Jiang, and Ma}]{tang2022piafusion}
Tang, L.; Yuan, J.; Zhang, H.; Jiang, X.; and Ma, J. 2022.
\newblock PIAFusion: A progressive infrared and visible image fusion network based on illumination aware.
\newblock \emph{Information Fusion}, 83: 79--92.

\bibitem[{Tirupal, Mohan, and Kumar(2021)}]{tirupal2021multimodal}
Tirupal, T.; Mohan, B.~C.; and Kumar, S.~S. 2021.
\newblock Multimodal medical image fusion techniques--a review.
\newblock \emph{Current Signal Transduction Therapy}, 16(2): 142--163.

\bibitem[{Wu et~al.(2025)Wu, Liu, Fu, Peng, Liu, Fan, and Liu}]{wu2025every}
Wu, G.; Liu, H.; Fu, H.; Peng, Y.; Liu, J.; Fan, X.; and Liu, R. 2025.
\newblock Every SAM drop counts: Embracing semantic priors for multi-modality image fusion and beyond.
\newblock In \emph{Proceedings of the Computer Vision and Pattern Recognition Conference}, 17882--17891.

\bibitem[{Xu et~al.(2026{\natexlab{a}})Xu, Li, Deng, Ma, and Liu}]{xu2026deno}
Xu, H.; Li, Y.; Deng, Y.; Ma, J.; and Liu, G. 2026{\natexlab{a}}.
\newblock Deno-if: Unsupervised noisy visible and infrared image fusion method.
\newblock \emph{Advances in Neural Information Processing Systems}, 38: 141649--141671.

\bibitem[{Xu et~al.(2020{\natexlab{a}})Xu, Ma, Jiang, Guo, and Ling}]{xu2020u2fusion}
Xu, H.; Ma, J.; Jiang, J.; Guo, X.; and Ling, H. 2020{\natexlab{a}}.
\newblock U2Fusion: A unified unsupervised image fusion network.
\newblock \emph{IEEE transactions on pattern analysis and machine intelligence}, 44(1): 502--518.

\bibitem[{Xu et~al.(2020{\natexlab{b}})Xu, Ma, Le, Jiang, and Guo}]{xu2020fusiondn}
Xu, H.; Ma, J.; Le, Z.; Jiang, J.; and Guo, X. 2020{\natexlab{b}}.
\newblock Fusiondn: A unified densely connected network for image fusion.
\newblock In \emph{Proceedings of the AAAI conference on artificial intelligence}, volume~34, 12484--12491.

\bibitem[{Xu, Ma, and Zhang(2020)}]{xu2020mef}
Xu, H.; Ma, J.; and Zhang, X.-P. 2020.
\newblock MEF-GAN: Multi-exposure image fusion via generative adversarial networks.
\newblock \emph{IEEE Transactions on Image Processing}, 29: 7203--7216.

\bibitem[{Xu et~al.(2025)Xu, Yi, Lu, Liu, and Ma}]{xu2025urfusion}
Xu, H.; Yi, X.; Lu, C.; Liu, G.; and Ma, J. 2025.
\newblock Urfusion: Unsupervised unified degradation-robust image fusion network.
\newblock \emph{IEEE Transactions on Image Processing}.

\bibitem[{Xu et~al.(2026{\natexlab{b}})Xu, Cao, Li, Hu, and Zhu}]{xu2026dream}
Xu, X.; Cao, B.; Li, D.; Hu, Q.; and Zhu, P. 2026{\natexlab{b}}.
\newblock Dream-IF: Dynamic Relative EnhAnceMent for Image Fusion.
\newblock In \emph{Proceedings of the AAAI Conference on Artificial Intelligence}, volume~40, 11442--11450.

\bibitem[{Xue et~al.(2024)Xue, Liu, Wang, He, and Zhuang}]{xue2024novel}
Xue, W.; Liu, Y.; Wang, F.; He, G.; and Zhuang, Y. 2024.
\newblock A novel teacher--student framework with degradation model for infrared--visible image fusion.
\newblock \emph{IEEE Transactions on Instrumentation and Measurement}, 73: 1--12.

\bibitem[{Yi et~al.(2024)Yi, Xu, Zhang, Tang, and Ma}]{yi2024text}
Yi, X.; Xu, H.; Zhang, H.; Tang, L.; and Ma, J. 2024.
\newblock Text-if: Leveraging semantic text guidance for degradation-aware and interactive image fusion.
\newblock In \emph{Proceedings of the IEEE/CVF Conference on Computer Vision and Pattern Recognition}, 27026--27035.

\bibitem[{Zhang et~al.(2025)Zhang, Cao, Zuo, Shao, and Ma}]{zhang2025omnifuse}
Zhang, H.; Cao, L.; Zuo, X.; Shao, Z.; and Ma, J. 2025.
\newblock Omnifuse: Composite degradation-robust image fusion with language-driven semantics.
\newblock \emph{IEEE Transactions on Pattern Analysis and Machine Intelligence}.

\bibitem[{Zhang et~al.(2021{\natexlab{a}})Zhang, Le, Shao, Xu, and Ma}]{zhang2021mff}
Zhang, H.; Le, Z.; Shao, Z.; Xu, H.; and Ma, J. 2021{\natexlab{a}}.
\newblock MFF-GAN: An unsupervised generative adversarial network with adaptive and gradient joint constraints for multi-focus image fusion.
\newblock \emph{Information Fusion}, 66: 40--53.

\bibitem[{Zhang and Ma(2021)}]{zhang2021sdnet}
Zhang, H.; and Ma, J. 2021.
\newblock SDNet: A versatile squeeze-and-decomposition network for real-time image fusion.
\newblock \emph{International Journal of Computer Vision}, 129(10): 2761--2785.

\bibitem[{Zhang et~al.(2021{\natexlab{b}})Zhang, Xu, Tian, Jiang, and Ma}]{zhang2021image}
Zhang, H.; Xu, H.; Tian, X.; Jiang, J.; and Ma, J. 2021{\natexlab{b}}.
\newblock Image fusion meets deep learning: A survey and perspective.
\newblock \emph{Information Fusion}, 76: 323--336.

\bibitem[{Zhang et~al.(2020{\natexlab{a}})Zhang, Xu, Xiao, Guo, and Ma}]{zhang2020rethinking}
Zhang, H.; Xu, H.; Xiao, Y.; Guo, X.; and Ma, J. 2020{\natexlab{a}}.
\newblock Rethinking the image fusion: A fast unified image fusion network based on proportional maintenance of gradient and intensity.
\newblock In \emph{Proceedings of the AAAI conference on artificial intelligence}, volume~34, 12797--12804.

\bibitem[{Zhang et~al.(2024)Zhang, Zuo, Jiang, Guo, and Ma}]{zhang2024mrfs}
Zhang, H.; Zuo, X.; Jiang, J.; Guo, C.; and Ma, J. 2024.
\newblock Mrfs: Mutually reinforcing image fusion and segmentation.
\newblock In \emph{Proceedings of the IEEE/CVF conference on computer vision and pattern recognition}, 26974--26983.

\bibitem[{Zhang et~al.(2020{\natexlab{b}})Zhang, Liao, Liu, Ma, Yang, and Xue}]{zhang2020real}
Zhang, J.; Liao, Q.; Liu, S.; Ma, H.; Yang, W.; and Xue, J.-H. 2020{\natexlab{b}}.
\newblock Real-MFF: A large realistic multi-focus image dataset with ground truth.
\newblock \emph{Pattern Recognition Letters}, 138: 370--377.

\bibitem[{Zhang(2021{\natexlab{a}})}]{zhang2021benchmarking}
Zhang, X. 2021{\natexlab{a}}.
\newblock Benchmarking and comparing multi-exposure image fusion algorithms.
\newblock \emph{Information Fusion}, 74: 111--131.

\bibitem[{Zhang(2021{\natexlab{b}})}]{zhang2021deep}
Zhang, X. 2021{\natexlab{b}}.
\newblock Deep learning-based multi-focus image fusion: A survey and a comparative study.
\newblock \emph{IEEE Transactions on Pattern Analysis and Machine Intelligence}, 44(9): 4819--4838.

\bibitem[{Zhang and Demiris(2023)}]{zhang2023visible}
Zhang, X.; and Demiris, Y. 2023.
\newblock Visible and infrared image fusion using deep learning.
\newblock \emph{IEEE Transactions on Pattern Analysis and Machine Intelligence}, 45(8): 10535--10554.

\bibitem[{Zhang et~al.(2020{\natexlab{c}})Zhang, Liu, Sun, Yan, Zhao, and Zhang}]{zhang2020ifcnn}
Zhang, Y.; Liu, Y.; Sun, P.; Yan, H.; Zhao, X.; and Zhang, L. 2020{\natexlab{c}}.
\newblock IFCNN: A general image fusion framework based on convolutional neural network.
\newblock \emph{Information Fusion}, 54: 99--118.

\bibitem[{Zhou et~al.(2024)Zhou, Zheng, He, Hong, and Chanussot}]{zhou2024probing}
Zhou, M.; Zheng, N.; He, X.; Hong, D.; and Chanussot, J. 2024.
\newblock Probing synergistic high-order interaction for multi-modal image fusion.
\newblock \emph{IEEE Transactions on Pattern Analysis and Machine Intelligence}, 47(2): 840--857.

\bibitem[{Zhu et~al.(2024)Zhu, Sun, Cao, and Hu}]{zhu2024task}
Zhu, P.; Sun, Y.; Cao, B.; and Hu, Q. 2024.
\newblock Task-customized mixture of adapters for general image fusion.
\newblock In \emph{Proceedings of the IEEE/CVF conference on computer vision and pattern recognition}, 7099--7108.

\end{thebibliography}


\end{document}

\maketitle
\subsection{More Implement details}
The framework supports three fusion tasks: VIF, MEF, and MFF. During training, we synthetically introduce diverse degradations, including defocus blur, downsampling, Gaussian noise, Gaussian blur, motion blur, haze, rain, snow, low illumination, overexposure, contrast reduction, and infrared nonuniformity, as illustrated in Fig.~\ref{fig:supp_sample}.
Each degradation is divided into five severity levels, which are sampled with probabilities ({0.05, 0.25, 0.40, 0.25, 0.05}). 
During inference, the diffusion process is performed using a single DDIM denoising step. We consider object detection and semantic segmentation as downstream applications. All three training stages use AdamW with a base learning rate of 1e-4, a weight decay of 0.01, and a OneCycle learning-rate schedule with cosine annealing. Each stage is trained for up to 20 epochs with a per-device batch size of 1.

% \subsection{Loss Function}
\subsection{Details of Fusion Losses}
\label{sec:fusion_loss}

Different fusion tasks follow distinct information aggregation principles. Therefore, instead of applying an identical objective to all tasks, we adopt task-specific fusion losses to supervise the shared fusion pathway. Given two source images $\mathbf{x}^1$ and $\mathbf{x}^2$ and the fused image $\mathbf{x}^f$, the objectives constrain structural similarity, intensity distribution, and gradient preservation according to the characteristics of each fusion task.

We first define the structural similarity loss as
\begin{equation}
\begin{aligned}
\mathcal{L}_{\mathrm{ssim}}
=&\,
\lambda_1
\left[1-\operatorname{SSIM}(\mathbf{x}^f,\mathbf{x}^1)\right] \\
&+
\lambda_2
\left[1-\operatorname{SSIM}(\mathbf{x}^f,\mathbf{x}^2)\right],
\end{aligned}
\label{eq:loss_ssim}
\end{equation}
where $\lambda_1=\lambda_2=0.5$.
Let $\nabla$ denote the Sobel gradient operator, while $\operatorname{max}(\cdot)$ and $\operatorname{mean}(\cdot)$ denote element-wise maximum and average operations, respectively. We further define $\operatorname{absmax}(\mathbf{A},\mathbf{B})$ as selecting the element with the larger absolute magnitude while retaining its original sign.

\paragraph{Visible--Infrared Fusion.}
For VIF, the fused image is expected to retain salient intensity responses and
strong structural details from both modalities. We therefore use
\begin{equation}
\mathcal{L}^{\mathrm{VIF}}_{\mathrm{fus}}
=
\mathcal{L}_{\mathrm{aux}}
+
\mathcal{L}_{\mathrm{ssim}}
+
\mathcal{L}_{\mathrm{max\text{-}int}}
+
\mathcal{L}_{\mathrm{max\text{-}grad}},
\label{eq:loss_vif}
\end{equation}
where
\begin{equation}
\mathcal{L}_{\mathrm{max\text{-}int}}
=
\frac{1}{HW}
\left\|
\mathbf{x}^f-
\operatorname{max}(\mathbf{x}^1,\mathbf{x}^2)
\right\|_1,
\label{eq:loss_max_int}
\end{equation}
and
\begin{equation}
\mathcal{L}_{\mathrm{max\text{-}grad}}
=
\frac{1}{HW}
\left\|
\nabla\mathbf{x}^f-
\operatorname{absmax}
(\nabla\mathbf{x}^1,\nabla\mathbf{x}^2)
\right\|_1.
\label{eq:loss_max_grad}
\end{equation}
The intensity term preserves prominent responses from either modality, whereas the gradient term transfers high-frequency structures without discarding gradient directions.

\paragraph{Multi-Exposure Fusion.}
MEF aims to produce a well-exposed image while retaining structural details from differently exposed inputs. Its objective is formulated as
\begin{equation}
\mathcal{L}^{\mathrm{MEF}}_{\mathrm{fus}}
=
\mathcal{L}_{\mathrm{aux}}
+
\mathcal{L}_{\mathrm{mefssim}}
+
\mathcal{L}_{\mathrm{avg\text{-}int}}
+
\mathcal{L}_{\mathrm{max\text{-}grad}},
\label{eq:loss_mef}
\end{equation}
where $\mathcal{L}_{\mathrm{mefssim}}$ is the MEF-specific structural similarity loss, and
\begin{equation}
\mathcal{L}_{\mathrm{avg\text{-}int}}
=
\frac{1}{HW}
\left\|
\mathbf{x}^f-
\operatorname{mean}(\mathbf{x}^1,\mathbf{x}^2)
\right\|_1.
\label{eq:loss_avg_int}
\end{equation}
The average-intensity constraint regularizes the global luminance of the fused image, while the gradient term preserves detailed structures across exposure levels.

\paragraph{Multi-Focus Fusion.}
For MFF, each local region should primarily inherit information from the better-focused source. Let $\mathbf{M}_i$ denote the focus mask associated with $\mathbf{I}_i$, where each mask identifies regions in which the corresponding source exhibits stronger local focus responses. The MFF objective is
\begin{equation}
\mathcal{L}^{\mathrm{MFF}}_{\mathrm{fus}}
=
\mathcal{L}_{\mathrm{aux}}
+
\mathcal{L}_{\mathrm{ssim}}
+
\mathcal{L}_{\mathrm{mask\text{-}int}}
+
\mathcal{L}_{\mathrm{mask\text{-}grad}},
\label{eq:loss_mff}
\end{equation}
with
\begin{equation}
\mathcal{L}_{\mathrm{mask\text{-}int}}
=
\sum_{i=1}^{2}
\left\|
\mathbf{M}_i\odot
(\mathbf{x}^f-\mathbf{I}_i)
\right\|_1,
\label{eq:loss_mask_int}
\end{equation}
and
\begin{equation}
\mathcal{L}_{\mathrm{mask\text{-}grad}}
=
\sum_{i=1}^{2}
\left\|
\mathbf{M}_i\odot
(\nabla\mathbf{x}^f-\nabla\mathbf{I}_i)
\right\|_1.
\label{eq:loss_mask_grad}
\end{equation}
These mask-guided terms encourage each fused region to preserve the intensity and edge information of its corresponding in-focus source.

\subsection{Quantitative Comparison of Multi-Task Fusion under Clean Inputs}
The quantitative comparisons of HP-UniIF with general-purpose and task-specific fusion methods under clean-input conditions are reported in Tables~\ref{tab:vif_clean}, \ref{tab:mef_clean}, and~\ref{tab:mff_clean}. For VIF, HP-UniIF achieves the best performance across all six metrics, with a slight improvement over DeFusion in MI and more pronounced gains in $Q^{\mathrm{AB/F}}$, VIFF, SCD, and SD. For MFF, HP-UniIF ranks first on four of the six metrics and second on the remaining two, demonstrating strong preservation of focused structures and complementary information despite joint training across heterogeneous fusion tasks. For MEF, HP-UniIF achieves the best MEF-SSIM and NIQE scores while ranking second in PSNR, MI, and VIFF, indicating a favorable balance between structural fidelity and perceptual quality. Overall, HP-UniIF ranks first or second on every reported metric across all three tasks. These results indicate that the shared framework effectively accommodates heterogeneous fusion objectives without an evident loss in task-specific performance. They also validate the effectiveness of TPM in adapting the shared diffusion prior to the distinct information-aggregation requirements of different fusion tasks.

\begin{table}[tbp]
\centering
\caption{Quantitative results of the VIF task on the LLVIP dataset. The best and second-best results are highlighted in \textbf{bold} and \underline{underlined}, respectively. 
% Underlined values indicate the best results among the compared general image fusion methods.
}
\label{tab:vif_clean}
\small
\setlength{\tabcolsep}{2.8pt}
\resizebox{\columnwidth}{!}{
\begin{tabular}{lcccccc}
\toprule
Method& EN$\uparrow$& $Q_{\mathrm{AB/F}}\uparrow$& MI$\uparrow$& VIFF$\uparrow$& SCD$\uparrow$& SD$\uparrow$\\
\midrule
U2Fusion (TPAMI'20) 
& 6.597& 0.427& 1.486& 0.479& 1.357 & 37.040\\
DeFusion (ECCV'22)
& 7.195& 0.421& \underline{1.954}& 0.582& 1.266 & 43.886\\
CCF (NeurIPS'24)
& 5.231& 0.480& 1.764& 0.590& 1.567& 44.054 \\
URFusion (TIP'25)
& 7.183& 0.499& 1.477& 0.529& 1.459& 51.094 \\
GIFNet (CVPR'25)
& 6.872& 0.394& 1.454& 0.462& 1.469& 43.267 \\
OmniFuse (TPAMI'25)
& 7.211& 0.281& 1.553& 0.580& 1.119& 43.192 \\
\midrule
% DRMF (ACM MM'24)
% & 7.061& 0.241& 1.147& 0.353& 1.070& 37.118 \\
% IASSF & 43.081 & 6.958 & 0.461 & \underline{\textit{1.990}} & 0.543 & 1.199 \\
TG-ECNet (ICML'25)
& 7.212& 0.470& 1.641& 0.555& 1.350 & 47.864\\
% RPFNet (ACM MM'25)
% & \underline{7.409} & \underline{0.519}& 1.719& \underline{0.655}& \underline{1.570}& \underline{47.809} \\
% SAGE (CVPR'25)
% & 6.817& 0.439& 1.618& 0.510& 1.479& 48.923 \\
TSDM-Fusion (TIM'26)
& 6.997& 0.474& 1.562& 0.571& 1.420& 41.441 \\
\midrule

% HP-UniIF (stage~1)
% & \textit{7.421}& \textit{0.591}& \textbf{2.025}& \textit{0.670}& \textit{1.618}& \textit{51.418} \\
HP-UniIF 
& \textbf{7.467}  & \textbf{0.605} & \textbf{1.965} & \textbf{0.693} & \textbf{1.619}& \textbf{52.916} \\

\bottomrule
\end{tabular}
}
\end{table}

\begin{table}[t]
\centering
\caption{
Quantitative comparison on the MFF task with clean images.
The best and second-best results are highlighted in
\textbf{bold} and \underline{underlined}, respectively.
All metrics are higher-is-better.
}
\label{tab:mff_clean}
\small
\setlength{\tabcolsep}{3.2pt}
\resizebox{\columnwidth}{!}{
\begin{tabular}{lcccccc}
\toprule
Method
& PSNR$\uparrow$
& $Q_{\mathrm{AB/F}}\uparrow$
& VIFF$\uparrow$
& NMI$\uparrow$
& FMI$\uparrow$
& $Q_W\uparrow$ \\
\midrule
U2Fusion
& 44.033
& 0.564
& \underline{0.989}
& 0.722
& 0.761
& \underline{0.842} \\
DeFusion
& \textbf{56.626}
& 0.479
& 0.948
& 0.800
& \underline{0.762}
& 0.694 \\
CCF
& 37.504
& 0.474
& 0.862
& \textbf{1.028}
& 0.747
& 0.767 \\
GIFNet
& 40.236
& 0.470
& 0.848
& 0.699
& \underline{0.762}
& 0.711 \\
OmniFuse
& 42.274
& 0.400
& 0.807
& 0.571
& 0.744
& 0.559 \\
URFusion
& 31.559
& 0.440
& 0.796
& 0.740
& 0.757
& 0.711 \\
\midrule
FusionDN
& 41.189
& 0.538
& 0.967
& 0.733
& 0.757
& 0.805 \\
PMGI
& 41.891
& 0.397
& 0.875
& 0.766
& 0.753
& 0.581 \\
FusionBooster
& 43.847
& \underline{0.568}
& 0.957
& 0.739
& 0.747
& 0.829 \\
\midrule
HP-UniIF
& \underline{53.111}
& \textbf{0.652}
& \textbf{1.080}
& \underline{0.929}
& \textbf{0.780}
& \textbf{0.887} \\
\bottomrule
\end{tabular}
}
\end{table}

\begin{table}[t]
\centering
\caption{
Quantitative comparison on the MEF task with clean images.
The best and second-best results are highlighted in
\textbf{bold} and \underline{underlined}, respectively.
All metrics are higher-is-better except NIQE.
}
\label{tab:mef_clean}
\small
\setlength{\tabcolsep}{3.2pt}
\resizebox{\columnwidth}{!}{
\begin{tabular}{lccccc}
\toprule
Method
& PSNR$\uparrow$
& MI$\uparrow$
& VIFF$\uparrow$
& MEF-SSIM$\uparrow$
& NIQE$\downarrow$ \\
\midrule
CCF
& 20.770
& \textbf{5.339}
& 0.977
& 0.860
& 5.174 \\
DeFusion
& 23.443
& 3.144
& 0.931
& 0.896
& \underline{4.124} \\
U2Fusion
& 21.552
& 3.849
& \textbf{1.197}
& \underline{0.918}
& 4.219 \\
GIFNet
& 21.498
& 3.903
& 1.061
& 0.862
& 4.319 \\
OmniFuse
& \textbf{25.364}
& 3.470
& 1.008
& 0.853
& 5.665 \\
URFusion
& 22.725
& 3.864
& 0.937
& 0.862
& 4.757 \\
\midrule
MEF-GAN
& 21.231
& 2.833
& 0.612
& 0.702
& 4.249 \\
Retinex-MEF
& 22.324
& 3.556
& 0.930
& 0.893
& 4.163 \\
\midrule
HP-UniIF
& \underline{24.628}
& \underline{5.332}
& \underline{1.184}
& \textbf{0.920}
& \textbf{3.787} \\
\bottomrule
\end{tabular}
}
\end{table}
\begin{figure*}[t]
\centering
\includegraphics[width=1\textwidth]{images/supp_degradation_sample.pdf} % Reduce the figure size so that it is slightly narrower than the column.
\caption{Representative sample of degradation in different fusion tasks.}
\label{fig:supp_sample}
\end{figure*}
\begin{figure*}[t]
\centering
\includegraphics[width=1\textwidth]{images/supp_det.pdf} % Reduce the figure size so that it is slightly narrower than the column.
\caption{Visualization of detection comparison.}
\label{fig:supp_det}
\end{figure*}
\begin{figure*}[t]
\centering
\includegraphics[width=1\textwidth]{images/supp_seg.pdf} % Reduce the figure size so that it is slightly narrower than the column.
\caption{Visualization of segmentation comparison.}
\label{fig:supp_seg}
\end{figure*}

\begin{figure*}[t]
\centering
\includegraphics[width=1\textwidth]{images/supp_VIF.pdf} % Reduce the figure size so that it is slightly narrower than the column.
\caption{Qualitative comparison of VIF.}
\label{fig:supp_vif}
\end{figure*}
\begin{figure*}[t]
\centering
\includegraphics[width=1\textwidth]{images/supp_MEF.pdf} % Reduce the figure size so that it is slightly narrower than the column.
\caption{Qualitative comparison of MEF.}
\label{fig:supp_mef}
\end{figure*}
\begin{figure*}[t]
\centering
\includegraphics[width=1\textwidth]{images/supp_MFF.pdf} % Reduce the figure size so that it is slightly narrower than the column.
\caption{Qualitative comparison of MFF.}
\label{fig:supp_mff}
\end{figure*}

% \bibliography{aaai2027,cited_references}